\documentclass[10pt,twocolumn,letterpaper]{article}

\usepackage[english]{babel}

\usepackage[T1]{fontenc}

\usepackage{stackengine}

\usepackage{array,ragged2e}

\usepackage{amsfonts} 

\usepackage{caption} 
\usepackage[a4paper,top=3cm,bottom=2cm,left=2cm,right=2cm, marginparwidth=1.75cm]{geometry}

\usepackage[backend=bibtex,style=numeric, maxnames=10]{biblatex}
\usepackage{makecell}

\usepackage{comment}
\usepackage{amsmath}
\usepackage{graphicx}
\usepackage[colorinlistoftodos]{todonotes}
\usepackage[colorlinks=true, allcolors=blue]{hyperref}

\title{\textbf{Exploring the sequence length bottleneck in the Transformer for Image Captioning}}

\usepackage{authblk}
\makeatletter
\renewcommand\AB@affilsepx{ - \protect\Affilfont}
\makeatother

\author[1]{Jia Cheng Hu}
\author[1]{Roberto Cavicchioli}
\author[1]{Alessandro Capotondi}
\affil[1]{\stackunder{University of Modena and Reggio Emilia}{\texttt{\{name.surname\}@unimore.it}}}

\date{}

\begin{document}
\maketitle

\selectlanguage{english}

\begin{abstract}
\textit{Most recent state of the art architectures rely on combinations and variations of three approaches: convolutional, recurrent and self-attentive methods. Our work attempts in laying the basis for a new research direction based upon the idea of modifying the sequence length. In order to do that, we propose a new method called ``Expansion Mechanism'' which transforms either dynamically or statically the input sequence into a new one featuring a different sequence length. Furthermore, we introduce a novel architecture that exploits such method and achieves competitive performances on the MS-COCO 2014 data set, yielding 134.6 and 131.4 CIDEr-D on the Karpathy test split in the ensemble and single model configuration respectively and 130 CIDEr-D in the official online evaluation server, despite being neither recurrent nor fully attentive. At the same time we address the efficiency aspect in our design and introduce a convenient training strategy suitable for most computational resources in contrast to the standard one. Source code is available at} \textnormal{\href{https://github.com/jchenghu/exploring}{https://github.com/jchenghu/exploring}}.

\end{abstract}

\section{Introduction}
The Image Captioning problem aims to provide natural language descriptions to images in an automated manner. 
It's a challenging task that requires both scenery comprehension as well as language understanding. Early approaches relied on statistical and graph based methods \cite{mitchell2012midge, kulkarni2013babytalk}, later with the advent of Neural Networks, models were able to extract richer visual features and deal with the text generation following an encoder-decoder structure \cite{anderson2018bottom, vinyals2015show, xu2015show, karpathy2015deep}, using CNN based architectures for the first part \cite{ren2015faster, szegedy2015going} and RNNs for the latter \cite{hochreiter1997long, cho2014learning}. Such framework were improved dramatically with the introduction of the Attention mechanism \cite{bahdanau2014neural, luong2015effective} which enabled a more effective connection between the two structures.

\begin{figure}[h]
\centering
\includegraphics[width=0.48\textwidth]{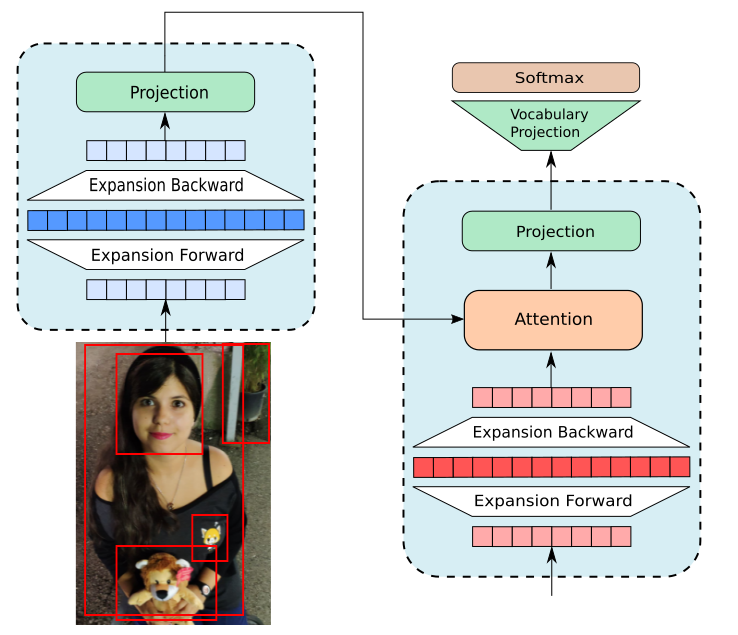}
\caption{\label{fig:expansion_mechanism_intro} Expansion mechanism. The expansion mechanism proposes an effective and efficient way for leveraging a different sequence length with respect to the input. It consists of a forward operation, in which the sequence length is modified, and a backward step in which the original length is retrieved back.}
\end{figure}

The Self-Attention method consists of applying the Attention to the encoding and decoding layers as well, introducing the first Fully Attentive model in the Transformer \cite{vaswani2017attention} which achieved the best performances in plenty of sequence modeling problems among stateless models \cite{gehring2017convolutional, vaswani2017attention}. Both recurrent and fully attentive blocks are often combined in hybrid architectures in Image Captioning and other fields \cite{chen2018best, pan2020x, huang2019attention, hao2019modeling} and perform on par if not better than the single approach ones.

In this work, we present an additional method in addition to the three currently most popular choices of convolutional, recurrent and attentive methods. In particular, we explore the possibility of improving performances by working with a different sequence length compared to the one provided by the input. At the same time, we address the problem of the computational expenses by designing formulation that mitigates the cost of a possible increased sequence length (the method can reduce the length as well). 

\textbf{Contributions.} The overall contributions of this work are the following:
\begin{itemize}
    \item[--] we introduce a new idea called Expansion Mechanism which allows to leverage a different length compared to the one provided in the input;
    \item[--] we present new layer based upon the main idea;
    \item[--] a new architecture called ExpansionNet that achieves competitive results despite not relying on any of the three currently most popular approaches;
    \item[--] we introduce a more efficient training strategy compared to the standard one defined by 30 epochs using a batch size of 10 in both Cross Entropy and CIDEr-D optimization. Our configuration instead enables a feasible training time for a much larger number of computational instances (e.g. our works focuses on the Nvidia Tesla K80).
\end{itemize}

\section{Related Works}

\subsection{Image Captioning}

Machine Learning field is constantly evolving and its applications are consequently following suit. Early Image Captioning system comprised of hand-crafted sentences combined with object detection outputs \cite{socher2010connecting, yao2010i2t}. With the advent of Deep Learning, the system split into a neural Encoder and Decoder, the first being responsible of extracting relevant visual features from the image, and the latter capable of generating the  description. Initially, the encoder consisted of a ConvNet trained over a large scale data set such as ImageNet \cite{deng2009imagenet} for the classification purpose and the features were fed into an LSTM \cite{hochreiter1997long, vinyals2015show, xu2015show}. Improvements were observed when such component was replaced with an object detector \cite{anderson2018bottom, ren2015faster} establishing a sequence modeling problem where the input is made of visual region features. Most decoders at the early stages were based upon RNNs \cite{wang2020show, anderson2018bottom, xu2015show, vinyals2015show} often in conjunction with the attention layer \cite{bahdanau2014neural, luong2015effective}, until fully attentive models \cite{vaswani2017attention} replaced them becoming the standard  ``de-facto'' approach in recent architectures \cite{herdade2019image, huang2019attention, cornia2020meshed, pan2020x}.

\subsection{Attentive models}

Attention mechanism was first introduced in Neural Machine Translation \cite{bahdanau2014neural, luong2015effective} and later successfully improved all existing RNN based encoder-decoder architectures in other research fields such as Image Captioning. The method consists of storing all recurrent states during the encoding phase and computing a weight distribution for all encoder's hidden state, by means of a softmax function, in each step of the decoding stage. 
The attention mechanism addresses one of the two main issues of the recurrent approach:
\begin{enumerate}
    \item the act of encapsulating the entire sequence into a single hidden vector represents a bottleneck on performance and leads to training difficulties;
    \item the sequential formulation inherently prevents GPUs from processing all elements simultaneously along the time direction. 
\end{enumerate}
Multiple proposal attempted in overcoming the second issue, for instance convolutional layers allow all the input tokens to be processed simultaneously without losing information on the position thanks to a moving context window \cite{bradbury2016quasi, bai2018empirical, gehring2017convolutional}, but these were later outperformed by the fully attentive ones \cite{vaswani2017attention, cornia2020meshed, sukhbaatar2019augmenting}. Fully attentive models extend the attention original purpose to the encoder and decoder layers as well in the so called self-attention layers. Several works successfully improved these architecture for captioning, \cite{herdade2019image} introduced geometrical awareness in the attention formula, whereas \cite{huang2019attention} modified the attentive layer by filtering noise in the query with a sigmoidal gate, finally \cite{pan2020x} achieved the best result by exploiting and extracting higher order features in the attentive block \cite{kim2018bilinear}.

Inspired by the fact that attention was originally designed for removing the bottleneck of a single hidden state in recurrent models, our work explores the question whether the target and source sequence length can be a bottleneck as well. We tackle this problem by proposing an alternative approach to the attention mechanism and addressing, at the same time, the efficiency aspect by designing computational friendly operations and focusing on stateless models in order to preserve the time-step wise parallel processing property.

\begin{figure}[h]
\centering
\includegraphics[width=0.45\textwidth]{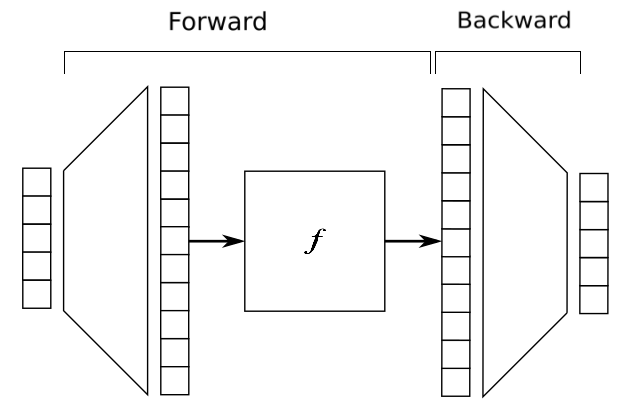}
\caption{\label{fig:expansion_mechanism_2} Expansion mechanism scheme in its essence.}
\end{figure}

\subsection{Other works}

To our knowledge, no other work explored the possibility of a performance bottleneck in the sequence length, however several works helped the development of the final result, for instance the Transformer \cite{vaswani2017attention} provided the main architectural inspiration, whereas BERT \cite{devlin2018bert}, in which special tokens are able to infer the missing words from a partially masked sentence, provides an insight of a possible reason why the method presented in the following sections works, as every expansion vector introduced by the expansion mechanism may be capable of extracting additional information compared to the original sequence.

\section{Method}

\subsection{Attention}
Given an input $X = \{x_1, x_2, ... , x_N\}$ and target sequence $Y = \{y_1, y_2, ... , y_M\}$ the attention mechanism \cite{bahdanau2014neural, luo2020better} refers to the practice of storing all the encoder's states $he_{i}, i \in \{1, ... , N\}$ and applying a weighted average by means of a softmax function with coefficients $\alpha_{i}, i \in \{1, ... ,N\}$ computed using a similarity function $f(hd_{j}, he_{i})$ according to a decoder's state $hd_{j}, j \in \{1, ... , M\}$:
\begin{equation}
\begin{aligned}
    e_{ij} &= f(hd_{j}, he_{i}) \\
    \alpha_{ij} &= Softmax(i, \{e_{1j}, ... , e_{Nj}\}) \\
    v_{j} &= \sum_{t}^{N} he_{i} \cdot \alpha_{ij}
\end{aligned}
\end{equation}
The method lied the basis of the self-attention \cite{vaswani2017attention} and all its variants such as
\cite{huang2019attention, pan2020x, wang2019r, gulati2020conformer}. The self-attention involves three input projections $K, V, Q \in \mathbb{R}^{N \times D}$ and it's formulated in matrix form as:
\begin{equation}
    \begin{aligned}
    s &= Softmax(\frac{Q K^{\top}}{\sqrt{D}})V \\
    out &= W^{out} s
    \end{aligned}
\end{equation}
in the single head formulation (we only showcase the single head case for the sake of simplicity). Additionally a transformation layer $W^{out} \in \mathbb{R}^{D \times D}$ is placed at the end for a total of 4 linear projections.

\subsection{Expansion Mechanism: Motivation}

The attention mechanism originally allowed the decoder to spread the input information along the whole collection of encoder's hidden vectors by means of the softmax function that inherently focuses on few elements only, a property after which the method is named. In contrast, the expansion we follow the idea of distributing the information over a different and arbitrary number of elements. As illustrated n Figure  \ref{fig:expansion_mechanism_2}, first, the sequence is transformed into another one featured by a new sequence length by means a ``Forward expansion''. Generic ope\-rations involving sequences in the new dimension can be done (represented by the symbol ``\textit{f}'' in the Figure), then the original length is retrieved back with an operation called ``Backward expansion''.

\subsection{Instance of Expansion implementation}

\begin{figure*}[ht!]
\centering
\includegraphics[width=1.0\textwidth]{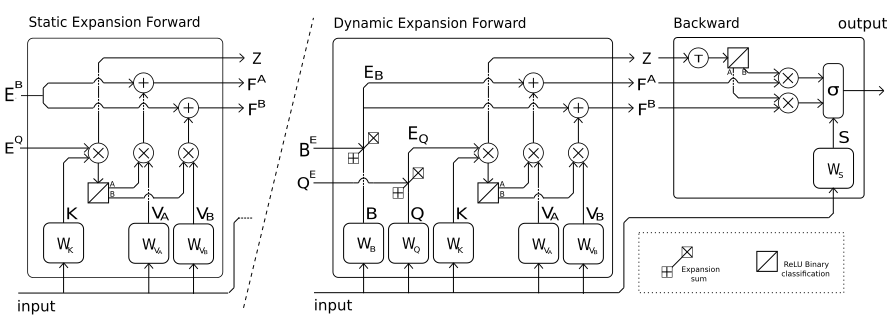}
\caption{\label{fig:dynamic_expansion.png} Illustration of the Expansion Layer in both static and dynamic expansion form. The diagonal dashed line indicates the differences between the two approaches in the forward step, whereas the backward block is shared by both methods. }
\end{figure*}

In this paragraph we propose one possible implementation of the expansion mechanism. In its essence, the idea consists of using a matrix multiplication for both Forward and Backward operations. We call such matrix ``Length Transformation Matrix'' denoted as \textit{Z} which it's shared in both steps.
The overall operation involves 4 up to 5 linear projections which produce the conditionings, keys, values and selectors denoted as $C, K, V_{1}, V_{2}, S \in \mathbb{R}^{T \times d_{model}}$ respectively. 

The first operation is the Forward expansion which consists of the vector generation phase and the processing phase. The first part is responsible of the generation of the so called expansion queries $Q^{E}$ and expansion bia\-ses $B^{E}$. Because of the sequential nature of the decoding step, we designed two kind of expansion, one suitable for the auto-regressive case and the other for the bidirectional one, differing only in the vector generation part:
\begin{itemize}
    \item[--] \textbf{Dynamic Expansion.} 
    In this case, given the input sequence of length $T$ and the expansion factor $N_{E}$,
    the expansion queries and biases $Q^{E}$, $B^{E}$ are defined simply by the sum between the input and two collection of parameters vectors $q^{E}_{i}, b^{E}_{i}$ $\in \mathbb{R}^{d_{model}}$, $i \in \{1,...,N_{E}\}$ according to the following formula:
    \begin{equation}
        \begin{aligned}
            e^{Q}_{ij} = c_{i} + q^{E}_{j} , \ \ c_{i} \in C, q^{E}_{j} \in Q^{E}\\ 
            e^{B}_{ij} = c_{i} + b^{E}_{j} , \ \ c_{i} \in C, b^{E}_{j} \in B^{E}\\
        \end{aligned}
    \end{equation}
    They are also represented in matrix notation as $E_{Q}, E_{B}$ $\in \mathbb{R}^{(T \cdot N_{E}) \times d_{model}}$. This type of expansion is designed for auto-regressive layers as well as bidirectional ones.
    \item[--]  \textbf{Static Expansion.}
    While in the previous case the coefficient $N_{E}$ is multiplied by the sequence length $T$. In case of static expansion, the expansion coefficient $N_{E}$ defines exactly the number of the expansion queries and biases and are defined as simple learnable vectors $q^{E}_{i}, b^{E}_{i}$ $\in \mathbb{R}^{d_{model}}$, $i \in \{1,...,N_{E}\}$
    \begin{equation}
        \begin{aligned}
            e^{Q}_{i} = q^{E}_{i} , \ \ q^{E}_{i} \in Q^{E}\\ 
            e^{B}_{i} = b^{E}_{i} , \ \ b^{E}_{i} \in B^{E}\\
        \end{aligned}
    \end{equation}
    represented also as $E_{Q}, E_{B} \in \mathbb{R}^{N_{E} \times d_{model}}$. This type of expansion does not involve the conditioning sequence $C$ and and can only be adopted in case of bidirectional processing.
\end{itemize}

\noindent
Given the definition of row-wise matrix normalization $\Phi: (X, \epsilon) \rightarrow Y$, $X,Y \in \mathbb{R}^{N_{1} \times N_{2}}, \epsilon \in \mathbb{R}_{>0}$:
\begin{equation}
    \Phi(X, \epsilon)_{ij} = \frac{x_{ij}}{\sum_{z}^{N_{2}}x_{iz} + \epsilon} 
    \label{eq:norm}
\end{equation}
which feasibility and numerical stability is guaranteed by the coefficient $\epsilon$, the remaining operations in the expansion consist of three steps:
\begin{enumerate}
    \item \textbf{Forward expansion:} Processing step. Given the expansion vectors generated according to one of the two criteria previously presented. In this step the dot product similarity is computed between the expanded queries and the keys:
    \begin{equation}
        Z =  \frac{E_{Q}K^{\top}}{\sqrt{d_{model}}}
    \label{eq:sim_forward}
    \end{equation}\\
    its result is fed into a ReLU function, normalized via Equation \ref{eq:norm} and the forward vectors are ultimately computed according to:
    \begin{equation}
    \begin{aligned}
    R^{fw}_{1} &= \Phi(ReLU(Z), \epsilon) \\
    R^{fw}_{2} &= \Phi(ReLU(-Z), \epsilon)
    \\ 
    F^{fw}_{1} &= R^{fw}_{1}V_{1} + E_{B} \\
    F^{fw}_{2} &= R^{fw}_{2}V_{2} + E_{B}
    \end{aligned}
    \end{equation}
    \item \textbf{Backward expansion}. This function performs similar operations compared to the previous step but it retrieves the original sequence length back by transposing the matrix in Equation \ref{eq:sim_forward}:
        \begin{equation}
        \begin{aligned}
        R^{bw}_{1} &= \Phi(ReLU(Z^{\top}), \epsilon) \\
        R^{bw}_{2} &= \Phi(ReLU(-Z^{\top}), \epsilon)
        \\ 
        B^{bw}_{1} &= R^{bw}_{1}F^{fw}_{1} \\
        B^{bw}_{2} &= R^{bw}_{2}F^{fw}_{2}
        \end{aligned}
        \end{equation}
    \item \textbf{Selection}.
        \begin{equation}
        out = \sigma(S) \odot B^{bw}_{1} + (1 - \sigma(S)) \odot B^{bw}_{2}
        \end{equation}
        $\sigma$ being the sigmoid gate function.
        
\end{enumerate}
During the forward transformation, the input information is distributed across a new sequence which length differs statically or dynamically with respect to the input, we suspect that the methods increase the quality of the input processing thanks to the additional information slot in case the new sequence length is greater than the previous one or the focus of the meaningful information over a smaller number of elements in the other way around.
In case of dynamic expansion, the extra elements resulted from the expansion of each single input, is hypothesized to be able to extract additional meaningful semantics from the whole sequence. Such information is then gathered back to each non expanded input in the backward step. Finally, the result is divided into two paths performing a sort of ``Neural Binary Classification'' for two purposes:
\begin{enumerate}
    \item we attempt in the creation of multiple specialized neural regions in which input signals can be properly conveyed \cite{kandel2000principles}
    implementing the active me\-mory in a different perspective, in contrast to \cite{sukhbaatar2019augmenting} which describes memory as a part of the network that affects each singular part of the input regardless the sequence as a whole (for e.g. the FeedForward); 
    \item it assures no information is lost during the Forward step. 
\end{enumerate}
Both static and dynamic expansion are depicted in Figure \ref{fig:dynamic_expansion.png} which visually showcases also the differences between the two.

\subsection{Expanded Encoder}

The encoder input consists of a collection of visual feature vectors $A = \{a_{1}, ..., a_{N}\}, \ a_{i} \in \mathbb{R}^{ d_{feature}}$ extracted by a Faster-RCNN \cite{ren2015faster}. Each feature is processed by a first linear transformation in order to adapt the dimension into the model's size $X = \{x_{1},..., x_{N}\}, \ x_{i} \in \mathbb{R}^{d_{model}}$, then it is refined by a stack of encoder layers, which consists of a expansion layer (which can be either static or dynamic) and a FeedForward layer, both wrapped in a residual connection and a pre-layer normalization structure:
\begin{equation}
    \begin{aligned}
        Y &= X + Exp^{ENC}(Norm(X)) \\
        O_{enc} &= Y + FF(Norm(Y))
    \end{aligned}
\end{equation}

\subsection{Expanded Decoder}

Given $Y = \{y_{1},..., y_{N}\}, \ y_{i} \in \mathbb{R}^{d_{model}}$ the decoded sequence. The expanded decoder (denoted as $Exp^{DEC})$ performs roughly the same operations with the only difference of the additional cross-layer component. However, in contrast to the encoder, only the dynamic expansion can be applied as the static one would violate the auto-regression condition: 
\begin{equation}
    \begin{aligned}
        Z &= Y + Exp^{DEC}(Norm(Y)) \\
        W &= Z + Cross\textnormal{-}Attention(Norm(Z), O_{enc}) \\
        O_{dec} &= W + FF(Norm(W))
    \end{aligned}
\end{equation}
The goal of the decoder, in the greedy configuration, consists of computing the conditional word log-probability step by step until the special termination \texttt{<eos>} token is the most likely token in the following formula:
\begin{equation}
    \begin{aligned}
        out_{t} &= f_{Net}(y_{1:t-1}, A; \theta) \\
        p(y_{t}|y_{1:t-1}, A) &= Softmax(out_{t})
    \end{aligned}
\end{equation}
where A is the collection of image features and $\theta$ the model's parameters.

\begin{figure*}[ht]
\centering
\includegraphics[width=0.8\textwidth]{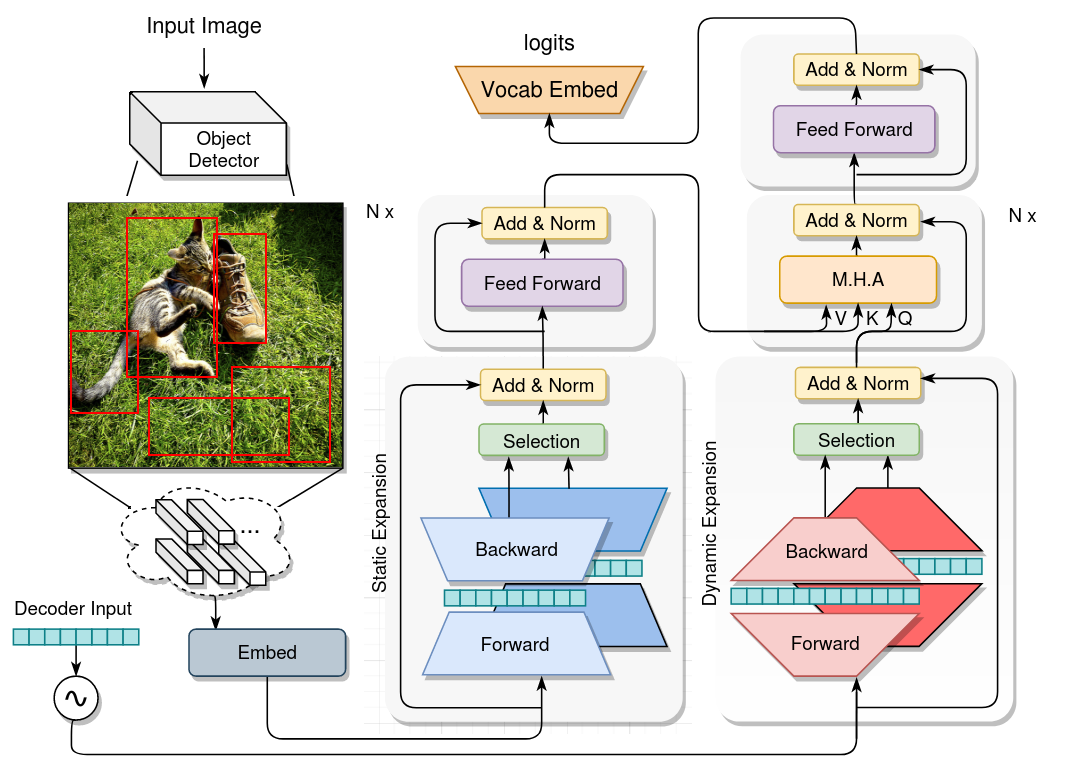}
\caption{\label{fig:intera_architettura_showcase.png} The Image Captioning system and ExpansionNet main components wrapped in skip connections.}
\end{figure*}

\subsection{Training objectives}

We follow the standard practice of pre-training the model using the Cross-Entropy loss:
\begin{equation}
    L_{XE}(\theta) = - \sum_{t}^{T} log( p_\theta(y^{*}_{t}|y^{*}_{1:t-1}))
\end{equation}
followed by the CIDEr-D \cite{vedantam2015cider} optimization in the Self-Critical Sequence Training \cite{rennie2017self}:
which approximate the gradient of the negative expected reward $L_{R}(\theta)=-\mathbf{E}_{y_{1:T~p_{\theta}}}[r(y_{1:T})]$ as:
\begin{equation}
        \nabla_{\theta} L_{R}(\theta) \approx -(r(y_{1:T}^s) - b) \nabla_\theta log \  p_\theta(y_{1:T}^s)
    \label{scst_loss}
\end{equation}
where $p_\theta$ is the output probability distribution, $r$ is the CIDEr-D reward, $y_{1:T}$ and $y^{*}_{1:T}$ are the sampled and ground truth caption respectively. The baseline $b$ is computed according to \cite{luo2020better} which is slightly more computational friendly with respect to the greedily decoded caption: 
\begin{equation}
    b(c_i) = \frac{1}{K} \sum_{j \neq i} R(c_j)   \ \ \ \ i = 1,...,K
\end{equation}
where K is set to 5.

\section{Experiments}

\subsection{Dataset}

We conduct the experiments on the popular Microsoft COCO benchmark \cite{lin2014microsoft} which consists a total of 123.287 images, 82.783 images for training, 40.504 and 40.775 images for validation and testing accordingly, ho\-wever it is split according to Karpathy \cite{karpathy2015deep} resulting in 113.287 training images and 5000 images for validation and an equal number for the offline testing. Each image is equipped with 5 reference captions and each caption is fed into a simple pre-processing pipeline that consist of lowering casing, a minimal punctuation filtering and the removal of words that do not occur at least 5 times, resulting in a vocabulary of size 10.000. The image region features \cite{anderson2018bottom} are extracted by the Faster R-CNN object detector \cite{ren2015faster} which is trained on the Visual Genome data set \cite{krishna2017visual} using the ResNet-101 backbone \cite{he2016deep}.

\subsection{Models}

Our model is named ``ExpansionNet\textsubscript{64,16}'' where 64 and 16 represent the static and dynamic expansion coefficients respectively. We compare our model with two baselines, first the base Transformer \cite{vaswani2017attention} (referred as ``Base'' in this work) with the number of encoding and decoding layer N=6 and  ``Base w/ AoA'' which is the same model but it's optimized according to \cite{huang2019attention}, no Feed-Forward and N=3. The latter differ from the original AoANet in that the LSTM is replaced by an auto-regressive self-attentive block for practical reasons (in our particular situation, stateful models require excessive machine time on K80 GPUs) and better comparison between stateless models. This additional baseline serves the purpose of better highlighting the impact of the training method presented in the next subsection. All baselines and ablation models use d\textsubscript{$model$}=512, d\textsubscript{$ff$}=2048, N=3 (except ``Base'') and n\textsubscript{$heads$}=8.


\subsection{Training details}

In order to enable this work to a wider number of resource instances, our experiments focus on the Nvidia Tesla K80 GPUs which are relatively dated compared to the recent architectures such networks are usually trained on for the COCO data set. For this reason our  configuration is mainly designed to produce results faster, without losing the validity of the experiments. In order to do so, we found a training pipeline which yields the same performance, if not better, on the baseline transformer compared to those observed in other works \cite{pan2020x, herdade2019image, huang2019attention} but in much less time, compared to the the popular practice of training both the cross entropy and reinforcement step for 30 epochs using a batch size 10 of (the standard configuration).
The latter is not suitable for the reference GPU because the small batch size, since the computational bottleneck lies in the gradient propagation step, prevents the training from leveraging multiple GPUs (see Table \ref{tab:train_speed_gpu}). Instead we splits the cross entropy training into two stages: 
\begin{enumerate}
    \item \textbf{Fast convergence step}: the model is trained with a batch size of 48 and a learning rate 2e-4 scheduled with warm-up of 10.000, annealed by 0.8 every 2 epochs. The number of epochs depends on the model, we stop when deterioration of the performances on the validation set is observed (approximately $\approx$ 12 epochs). 
    \item \textbf{Fine-tuning step}: in order to fill the gap of the regularization property with respect to the standard configuration, we complete the pre-training step with one additional epochs using batch size 10 and fixed learning rate 5e-6. This step is optional, in particular, it's applied only when the model's performance improves from this step, which is always the case in baselines and ablative models.
\end{enumerate}
The Self Critical Learning step is trained for approximately 35 epochs using the same stopping criteria, a batch size of 32 and the same learning rate algorithm starting from an initial value of 1e-4. 
\begin{table}[ht!]
  \caption{Time comparison of Cross-Entropy and SCST training using the base model. ``$\times$'' symbol means out of memory (we do not consider accumulations to not compromise the measurements).}
  \footnotesize
  \center
  \begin{tabular}{| c | c | c | c | c | c |} 
 \hline
 Model & N & GPUs & Ours & Standard & Gain \\
 \hline
 \multicolumn{6}{|c|}{Cross Entropy}\\
 \hline
 Base & 6 & 1 & 20 h & 90 h & 4.5 \\
 \hline
 Base & 6 & 4 & 9 h & 90 h & 10 \\
 \hline
 \multicolumn{6}{|c|}{SCST Training }\\
 \hline
 Base & 6 & 1 & $\times$ & 210 h &  {/} \\
 \hline
 Base & 6 & 4 & 58 h & 210 h & 3.6 \\
 \hline
 \end{tabular}
  \label{tab:train_speed_gpu}
\end{table}
The benefits of such configuration can be observed in Table \ref{tab:train_speed_gpu} and Table \ref{tab:legitness_train_conf} demonstrates its vali\-dity but it most importantly shows that even though cross entropy training results are heavily penalized, it ultimately little affects the model performances at the end of the reinforcement learning process.

\begin{table*}[htb!]
  \centering
  \small
  \caption{Our models and baselines scores comparison on the test set using using beam size 2 in the Cross-Entropy stage and beam size 5 in the CIDEr-D optimization one.}
  \begin{tabular}{| c | c | c | c | c | c | c | c | c | c |  c | c | c | c | c | c | c | c |} 
 \hline
 \multicolumn{1}{|c|}{} & \multicolumn{6}{c|}{Cross Entropy} & \multicolumn{6}{c|}{CIDEr-D optimization} \\
 \hline
 Model & B1 & B4 & M & R & C & S & B1 & B4 & M & R & C & S\\
 \hline
 Base \cite{pan2020x} & 76.1 & 34.0 & 27.6 & 56.2 & 113.3 & 21.0 & 80.2 & 38.6 & 28.8 & 58.5 & 128.3 & 22.6 \\
 \hline
 Base (ours)& 73.7 & 33.6 & 28.0 & 56.0 & 110.4 & 21.0 & 80.2 & 37.7 & 28.5 & 58.2 & 128.2 & 22.4 \\
 \hline
 Base w/ AoA & 75.2 & 35.2 & 28.5 & 56.8 & 115.4 & 21.5 & 80.3 & 38.2 & 28.6 & 58.3 & 129.2 & 22.4 \\
 \hline
 ExpansionNet\textsubscript{\raisebox{-1pt}{\rlap{64,1}}}\textsuperscript{\raisebox{1pt}{Ablation}}
 & 75.6 & 35.4 & 28.5 & 57.0 & 116.2 & 21.6 & 80.4 & 38.5 & 28.8 & 58.6 & 129.3 & 22.4 \\
 \hline 
 ExpansionNet\textsubscript{\raisebox{-1pt}{\rlap{64,16}}}\textsuperscript{\raisebox{1pt}{Ablation}}
 & 75.7 & 35.4 & 28.5 & 56.8 & 117.0 & 21.7 & 80.7 & 38.5 & 28.9 & 58.7 & 130.2 & 22.6 \\
 \hline
 \end{tabular}
  \label{tab:legitness_train_conf}
\end{table*}

\begin{table*}[htb!]
  \centering
  \small
  \caption{Ablation study evaluating on the test set  in the Cross-Entropy training phase using beam size 2.}
  \begin{tabular}{| c | c | c | c | c | c | c | c | c | c |} 
 \hline
 Encoder & Decoder & B1 & B2 & B3 & B4 & M & R & C & S  \\
 \hline
 Base & Base & 73.7 & 57.5 & 44.0 & 33.6 & 28.0 & 56.0 & 110.4 & 21.0\\
 \hline
 Static Exp$_{N_{E}=64}$ & Base & 74.8 & 58.7 & 45.0 & 34.3 & 28.2 & 56.3 & 114.1 & 21.4 \\
 \hline
 Dynamic Exp$_{N_{E}=1}$ & Base & 75.0 & 58.7 & 45.0 & 34.4 & 28.3 & 56.4 & 114.2 & 21.5 \\
 \hline
  Dynamic Exp$_{N_{E}=8}$ & Base & 74.9 & 58.9 & 45.3 & 34.6 & 28.3 & 56.4 & 114.4 & 21.5 \\
 \hline
  Dynamic Exp$_{N_{E}=16}$ & Base & 74.8 & 58.6 & 45.0 & 34.5 & 28.3 & 56.3 & 114.4 & 21.6 \\
 \hline
  Static Exp$_{N_{E}=64}$ & Dynamic Exp$_{N_{E}=1}$ & 75.6 & 59.7 & 46.0 & 35.4 & 28.5 & 57.0 & 116.2 & 21.6 \\
 \hline
 Static Exp$_{N_{E}=64}$ & Dynamic Exp$_{N_{E}=8}$ & 75.8 & 59.9 & 46.3 & 35.6 & 28.4 & 56.9 & 116.5 & 21.7 \\
 \hline
   Static Exp$_{N_{E}=64}$ & Dynamic Exp$_{N_{E}=16}$ & 75.7 & 59.8 & 46.2 & 35.4 & 28.5 & 56.8 & 117.0 & 21.7 \\
  \hline
 \end{tabular}
 \label{tab:ablation}
\end{table*}

\begin{figure}[ht!]
\centering
\includegraphics[width=0.48\textwidth]{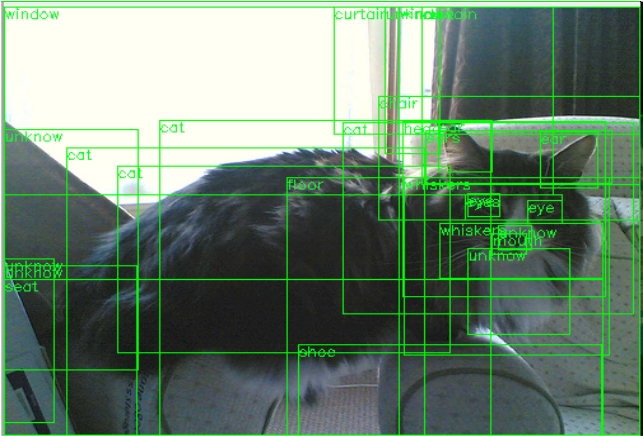}
\caption{\label{fig:img_bounding.png} Elements in the coco images are typically detected multiple times by the object detectors, in the present image for instance, the class ``cat'' is detected four times. Such abundance it's assumed to be the reason why different expansion coefficients in the encoder yields only slightly different performances.}
\end{figure}

\subsection{Ablation Study}

In order to compare the performances of the expansion mechanism with respect to the baseline we replace the encoder and decoder in the Base model with our Expansion counterparts and evaluate several settings of expansion
coefficient $N_{E}$. \\

\medskip
\noindent
\textbf{Expanded Encoder}: In this module both static and dynamic expansions are implemented in the encoder using different parameter sets, while the Base decoder is left as it is. In Table \ref{tab:ablation} it can be observed that the new model performs better than the baseline in all instances suggesting the benefits of the expansion. However, compared to the decoder's case, no much difference can be seen between dynamic and static expansion and much less expansion is needed. We hypothesize that it is due to the abundant number of detected objects provided by the backbone network. In fact, state of the art detectors often provide multiple bounding boxes over the same region (an example is provided in Figure \ref{fig:img_bounding.png}). Similarly not much difference is observed in case of dynamic expansion, for this reason, in the remaining experiments we choose a static coefficient of 64 which uses slightly less memory and computational resources.\\

\medskip
\noindent
\textbf{Expanded Decoder}: As previously described, we fix the encoder expansion coefficient to a constant (Static Exp$_{N_{E}=64}$) and perform experiments with different choices of decoder configuration. The decoder's expansion provides another boost to the performances which increases proportionally to the expansion coefficient. But this time in a more convincing manner compared to the encoder's case introducing a difference of 0.8 CIDEr-D between the smallest and biggest score, in addition to the 2 points increase with respect to the baseline decoder. The two results are reflected also in the reinforcement step, as illustrated in Table \ref{tab:ablation} (named ExpansionNet\textsubscript{\raisebox{-1pt}{\rlap{64,1}}}\textsuperscript{\raisebox{1pt}{Ablation}} and ExpansionNet\textsubscript{\raisebox{-1pt}{\rlap{64,16}}}\textsuperscript{\raisebox{1pt}{Ablation}} respectively), thus proving the effectiveness of the expansion module.

\subsection{Comparison}

\begin{table*}[htb!]
  \centering
  \footnotesize
  \caption{Offline Comparison of state of the art models trained on ResNet-101 visual features.}
  \begin{tabular}{| c | c | c | c | c | c | c | c | c | c | c | c | c | c | c |} 
 \hline
 \multicolumn{1}{|c|}{} & \multicolumn{6}{c|}{Cross-Entropy} & \multicolumn{6}{c|}{CIDEr-D optimization}\\
 \hline
 Model & B@1 & B@4 & M & R & C & S & B@1 & B@4 & M & R & C & S \\
 \hline
 \multicolumn{13}{|c|}{Single model}\\
 \hline
 LSTM \cite{vinyals2015show}& - & 29.6 & 25.2 & 52.6 & 94 & - & - & 31.9 & 25.5 & 54.3 & 106.3 & -\\
 \hline
 SCST\textsubscript{Att2All} \cite{rennie2017self} & - & 30.0 & 25.9 & 53.4 & 99.4 & - & - & 34.2 & 26.7 & 55.7 & 114.0 & -\\
 \hline
 Up-Down \cite{anderson2018bottom} & 77.2 & 36.2 & 27.0 & 56.4 & 113.5 & 20.3 & 79.8 & 36.3 & 27.7 & 56.9 & 120.1 & 21.4\\
 \hline
 GCN-LSTM \cite{yao2018exploring} & 77.3 & 36.8 & 27.9 & 57.0 & 116.3 & 20.9 & 80.5 & 38.2 & 28.5 & 58.3 & 127.6 & 22.0\\
 \hline
 SGAE \cite{yang2019auto} & - & - & - & - & - & - & 80.8 & 38.4 & 28.4 & 58.6 & 127.8 & 22.1\\
 \hline
 Base (ours) & 73.7 & 33.6 & 28.0 & 56.0 & 110.4 & 21.0 & 80.2 & 37.7 & 28.5 & 58.2 & 128.2 & 22.4\\
 \hline
  ObjRel. Base \cite{herdade2019image} & 76.6 & 35.5 & 28.0 & 56.6 & 115.4 & 21.2 & 80.5 & 38.6 & 28.7 & 58.4 & 128.3 & 22.6\\
 \hline
 AoANet \cite{huang2019attention}& 77.4 & 37.2 & 28.4 & 57.5 & 119.8 & 21.3 & 80.2 & 38.9 & 29.2 & 58.8 & 129.8 & 22.4\\
 \Xhline{1.5\arrayrulewidth}
 ExpansionNet\textsubscript{\raisebox{-1pt}{\rlap{64,16}}}\textsuperscript{\raisebox{1pt}{Ablation}} & 75.7 & 35.4 & 28.5 & 56.9 & 117.0 & 21.7 & 80.7 & 38.5 & 28.9 & 58.7 & 130.2 & 22.6 \\
 \hline
 ExpansionNet\textsubscript{64,16} & 76.1 & 35.6 & 28.7 & 57.1 & 118.7 & 21.8 & 80.9 & 38.9 & 29.1 & 58.8 & 131.4 & 22.9 \\
 \hline
 \multicolumn{13}{|c|}{Ensemble model}\\
 \hline
 SCST\textsuperscript{$\sum$}\textsubscript{Att2All} \cite{rennie2017self} & - & 32.8 & 26.7 & 55.1 & 106.5 & - & - & 35.4 & 27.1 & 56.6 & 117.5 & -\\
 \hline
 GCN-LSTM\textsuperscript{$\sum$}\cite{yao2018exploring} & 77.4 & 37.1 & 28.1 & 57.2 & 117.1 & 21.1 & 80.9 & 38.3 & 28.6 & 58.5 & 128.7 & 22.1\\
 \hline
 SGAE\textsuperscript{$\sum$}\cite{yang2019auto} & - & - & - & - & - & - & 81.0 & 39.0 & 28.4 & 58.9 & 129.1 & 22.2\\
 \hline
 AoANet\textsuperscript{$\sum$}\cite{huang2019attention} & 78.7 & 38.1 & 28.5 & 58.2 & 122.7 & 21.7 & 81.6 & 40.2 & 29.3 & 59.4 & 132.0 & 22.8\\
 \Xhline{1.5\arrayrulewidth}
 ExpansionNet\textsuperscript{$\sum_2$}\textsubscript{64,16} & 76.7 & 36.4 & 28.9 & 57.4 & 120.7 & 21.9 & 81.3 & 39.6 & 29.2 & 59.1 & 133.0 & 23.0 \\
 \hline
 ExpansionNet\textsuperscript{$\sum$}\textsubscript{64,16} & 77.2 & 37.0 & 28.7 & 57.4 & 120.5 & 21.7 & 81.5 & 40.1 & 29.4 & 59.4 & 134.6 & 23.1 \\
 \hline
 \end{tabular}
  \label{tab:offline_table_eval}
\end{table*} 

\begin{table*}[htb!]
  \footnotesize
  \center
  \caption{Online Evaluation of state of the art models trained on ResNet-101 visual features.}
  \begin{tabular}{| c | c | c | c | c | c | c | c | c | c | c | c | c | c | c |} 
 \hline
 Model & \multicolumn{2}{c|}{B1} & \multicolumn{2}{c|}{B2} &\multicolumn{2}{c|}{ B3} & \multicolumn{2}{c|}{B4} & \multicolumn{2}{c|}{METEOR} & \multicolumn{2}{c|}{ROUGE-L} & \multicolumn{2}{c|}{CIDEr-D}\\
 \hline
 {} & c5 & c40 & c5 & c40 & c5 & c40 & c5 & c40 & c5 & c40 & c5 & c40 & c5 & c40\\
 \hline
 SCST\textsuperscript{$\sum$} \cite{rennie2017self} & 78.1 & 93.7 & 61.9 & 86.0 & 47.0 & 75.9 & 35.2 & 64.5 & 27.0 & 35.5 & 56.3 & 70.7 & 114.7 & 116.0 \\
 \hline
 Up-Down\textsuperscript{$\sum$} \cite{anderson2018bottom}   & 80.2 & 95.2 & 64.1 & 88.8 & 49.1 & 79.4 & 36.9 & 68.5 & 27.6 & 36.7 & 57.1 & 72.4 & 117.9 & 120.5\\
 \hline
 GCN-LSTM\textsuperscript{$\sum$} \cite{yao2018exploring} & - & - & 
 65.5 & 89.3 & 50.8 & 80.3 & 38.7 & 69.7 & 28.5 & 37.6 & 58.5 &
 73.4 & 125.3 & 126.5 \\
 \hline
 SGAE\textsuperscript{$\sum$} \cite{yang2019auto}& 81.0 & 95.3 &
 65.6 & 89.5 & 50.7 & 80.4 & 38.5 & 69.7 & 28.2 & 37.2 & 58.6 & 73.6 & 123.8 & 126.5 \\
 \hline
 AoANet\textsuperscript{$\sum$} \cite{huang2019attention}& 81.0 & 95.0 &
 65.8 & 89.6 & 51.4 & 81.3 & 39.4 & 71.2 & 29.1 & 38.5 & 58.9 & 74.5 & 126.9 & 129.6 \\
 \hline
 X-Transformer\textsuperscript{$\sum$} \cite{pan2020x} & 81.3 & 95.4 & 66.3 & 90.0 & 51.9 & 81.7 & 39.9 & 71.8 & 29.5 & 39.0 & 59.3 & 74.9 & 129.3 & 131.4 \\
 \Xhline{1.5\arrayrulewidth}
 ExpansionNet\textsuperscript{$\sum$}\textsubscript{64,16} & 80.9 & 95.2 & 65.7 & 89.7 & 51.1 & 81.1 & 39.0 & 70.6 & 29.0 & 38.2 & 58.8 & 73.9 & 127.8 & 130.0  \\
 \hline
 \end{tabular}
  \label{tab:online_table_eval}
\end{table*} 

\begin{figure*}[ht]
\centering
\includegraphics[width=0.8\textwidth]{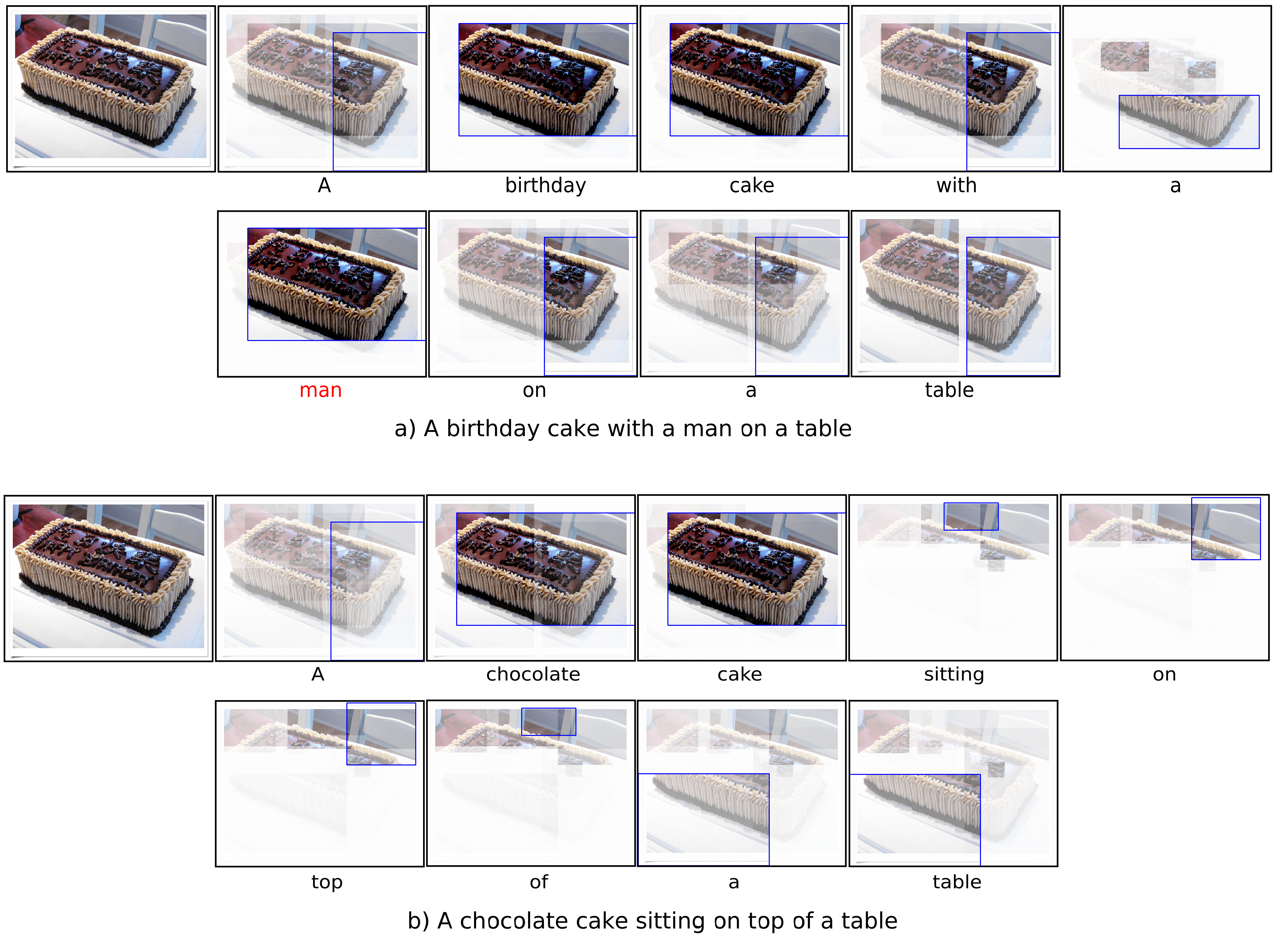}
\caption{\label{fig:attention_visualization.png} Visualization of attention of the baseline model (part a) and ExpansionNet (part b).}
\end{figure*}

\subsubsection{Offline Evaluation}

The performance comparison in the MS-COCO 2014 testing set between our proposed model and others are summarized in Table \ref{tab:offline_table_eval}. Regarding the most recent models, only those which backbone relies on the ResNet-101 are considered. Show and Tell \cite{xu2015show} consists of a fully convolutive encoder where the input is a single visual feature vector, and the LSTM \cite{hochreiter1997long} is responsible of the decoding stage, Up-Down \cite{anderson2018bottom} instead, splits the visual input into a set of features representing salient boxes in the original image generated by an object detector. SCST \cite{rennie2017self} differs from the first presented model in this section, using an alternative pathing of attention signals in the recurrent model. GCN-LSTM \cite{yao2018exploring} and SGAE \cite{yang2019auto} achieve better performances by exploiting additional semantic information like scene graphs. Finally the ObjRel. Transf \cite{herdade2019image} and AoANet \cite{huang2019attention} are originated from our same baseline which combines the Faster-RCNN \cite{ren2015faster} in the encoder and the Transformer \cite{vaswani2017attention} in both encoder and decoder. The first provides the bounding boxes geometrical information in the attention formula and the latter outperforms all the previous model by refining the attentive parts and using an additional recurrent component in the decoder. 
We call our model ExpansionNet, and present two instances of it, one named ExpansionNet\textsubscript{\raisebox{-1pt}{\rlap{64,16}}}\textsuperscript{\raisebox{1pt}{Ablation}} that is configured for the ablation analysis and another one we simply refer as
ExpansionNet\textsubscript{64,16} which is further optimized and fine-tuned for better performances inspired by the works \cite{pan2020x, liu2019variance}, in particular, it consists of additional linear projections and it is trained for a bigger number of epochs in the cross entropy loss. Our model does not achieve the best results but in the SPICE score in case of Cross-Entropy due to our particular training phase that was designed for yielding good results in much less epochs, overcoming the computational limitations. However, at the end of the reinforcement learning it outperforms all the previous models on the BLEU1, CIDEr-D and SPICE metrics and it's on par with the AoANet in the BLEU4 and ROUGE score. The improvements are reflected also in case of an ensemble model generated using different initialization seeds and it outperforms all the other reported scores in case of ResNet-101 features using either two or four instances instances referred as ExpansionNet\textsubscript{64,16}$^{\sum_2}$ and ExpansionNet\textsubscript{64,16}$^{\sum}$ respectively. 

\subsubsection{Online Evaluation}

The model scores one and half CIDEr-D below the best performing model at the time this project was developed (years 2020-2021) on the online testing server (Table \ref{tab:online_table_eval}), which is unfortunately lower than what offline scores forecasted. 
Nonetheless it achieves a very competitive good result for a not fully attentive model and that can be further improved in a plenty of ways, for example, by applying all the recent advances in Image Captioning \cite{huang2019attention, pan2020x} to the cross-connection part, and adopting a more performance focused training configuration.

\vspace{-0.2cm}
\subsection{Computational Perspective}

The expansion mechanism does not introduce much additional computational cost as the GPU performs a single tensor multiplication in the forward step regardless of any given expansion coefficient $N_E$ and the backward step ensures the cost is not propagated to the remaining and more expensive parts of the network. In particular, the overall design involves limited size matrix multiplications, in particular
at no point the expanded sequence is fed into a fully connected layer (such as the Feed Forward layer). This allows our model to exploit the advantages of a longer input sequence without the computational burden of the scenario in which the input sequence length is actually increased throghout the whole network. In fact, even in the case of the highest coefficients configuration (Static Exp $N_E = 64$ and Dynamic Exp $N_E = 16$) the training time cost in the ablation model is barely doubled in the least efficient computational case of a single Tesla K80.

\subsection{Qualitative Analysis}

Tables \ref{tab:example_captions_1} and \ref{tab:example_captions_2} showcase some examples of image captioning results of the stateless models coupled with the respective ground truth sentences. Selected examples ranges from simple scenarios such as image 1, 2, 3, 5, 8, to images with richer number elements and details such as 4, 6, 7. Our model seem to yield a more correct description compared to the baselines like in image number 1, 2, 3, 5, 6. Moreover, in case all descriptions are correct, our model seems to provide a more descriptive prediction such as image number 8. 

Baselines sometimes predict very unlikely situations such as example 2, 3 and 5 and Figure \ref{fig:attention_visualization.png} further showcases this problem providing, at the same time, an example of attention visualization.

\begin{table*}[htb]
 \centering
  \footnotesize
  \caption{Examples of predicted captions using beam search (width 5), the respective ground truth and comparison between baselines and our models}.
  \begin{tabular}[t]{ p{0.3cm}  @{}c@{} | >{\RaggedRight\arraybackslash}p{4.8cm} | >{\RaggedRight\arraybackslash}p{4.8cm} } 
 
 {} & Image & Networks prediction & Ground Truth \\

  \hline
 1 &
 \raisebox{-0.9\totalheight}{
 \includegraphics[height=0.3\textwidth, width=0.3\textwidth]{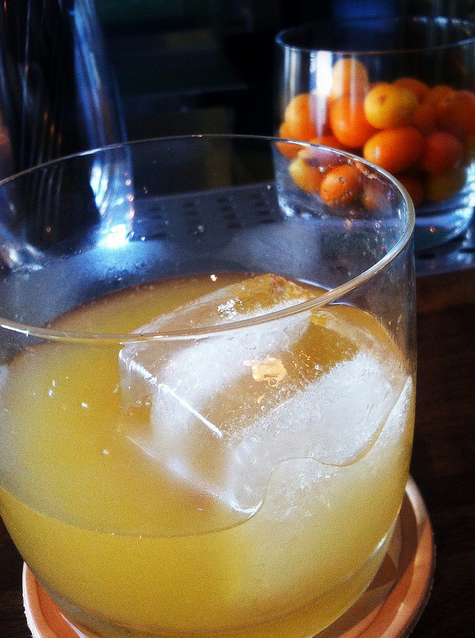}}
& {Base: A glass bowl of ice cream on a table. \leavevmode\newline \leavevmode\newline
Base w/ AoA: A blender with oranges and a man on a plate. \leavevmode\newline \leavevmode\newline
ExpansionNet\textsubscript{\raisebox{-1pt}{\rlap{64,16}}}\textsuperscript{\raisebox{1pt}{Ablation}}: A glass of soup with a bowl of oranges. \leavevmode\newline \leavevmode\newline
ExpansionNet\textsubscript{64,16}: A glass of orange juice next to a plate of oranges.
} & {Gt1: There is some type of drink in a small cup with ice. \newline Gt2: A drink in a glass with an ice cube. \newline Gt3: Clear glass of orange beverage with ice cube. \newline Gt4: A glass of beverage served chilled with ice. \newline Gt5: A yellow beverage with a large ice cube in a glass.} \\

 \hline
2 &
 \raisebox{-0.9\totalheight}{
 \includegraphics[height=0.3\textwidth,width=0.3\textwidth]{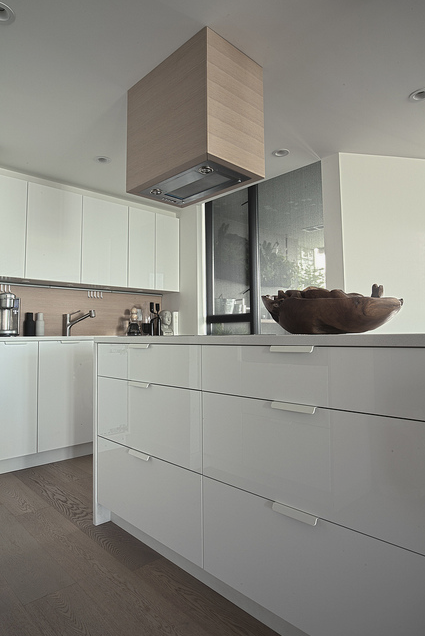}}
& {Base: A cat is sitting on top of a kitchen counter. \leavevmode\newline \leavevmode\newline
Base w/ AoA: A kitchen with two windows and a cow on a counter. \leavevmode\newline \leavevmode\newline
ExpansionNet\textsubscript{\raisebox{-1pt}{\rlap{64,16}}}\textsuperscript{\raisebox{1pt}{Ablation}}: A kitchen with white cabinets and a sink.
\leavevmode\newline \leavevmode\newline
ExpansionNet\textsubscript{64,16}: A kitchen with white cabinets and a bowl on a counter.
} & {Gt1: A bowl of food sitting on top of a white kitchen counter. \newline Gt2: A large number of white cabinets in a kitchen. \newline Gt3: A white modern kitchen is on display in this photo. \newline Gt4: A spotless white kitchen with some sort of platter on the counter. \newline Gt5: A white kitchen counter with a big, brown bowl on it.} \\

 \hline
 3 &
 \raisebox{-0.9\totalheight}{
 \includegraphics[height=0.3\textwidth,width=0.3\textwidth]{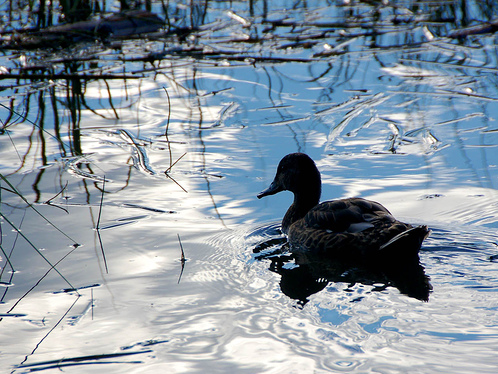}}
& {Base: A duck is swimming in the snow. \leavevmode\newline \leavevmode\newline
Base w/ AoA: Two ducks swim\-ming in the snow on a table. \leavevmode\newline \leavevmode\newline
ExpansionNet\textsubscript{\raisebox{-1pt}{\rlap{64,16}}}\textsuperscript{\raisebox{1pt}{Ablation}}: A black duck swimming in the water.
\leavevmode\newline \leavevmode\newline
ExpansionNet\textsubscript{64,16}: A black duck swimming in the water.
} & 
{Gt1: A brown duck floats by itself on the water. \newline Gt2: A duck swimming in the water, in a pond. \newline Gt3: The duck is in the  moving water. \newline Gt4: A duck is swimming in the pond to the next destination. \newline Gt5: A duck is swimming alone is a pond.} \\

 \hline
 4 &
 \raisebox{-0.9\totalheight}{
 \includegraphics[height=0.3\textwidth,width=0.3\textwidth]{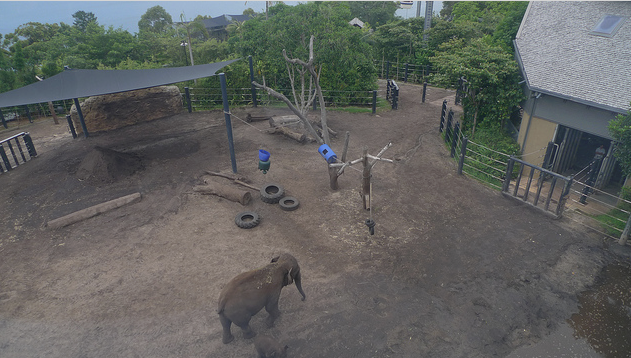}}
& {Base: A baby elephant walking in a dirt field. \leavevmode\newline \leavevmode\newline
Base w/ AoA: A baby elephant standing next to a group of elephants. \leavevmode\newline \leavevmode\newline
ExpansionNet\textsubscript{\raisebox{-1pt}{\rlap{64,16}}}\textsuperscript{\raisebox{1pt}{Ablation}}: A baby elephant walking in front of a zoo.
\leavevmode\newline \leavevmode\newline
ExpansionNet\textsubscript{64,16}: A baby elephant walking in a zoo.
} & {Gt1: A small elephant standing next to a house. \newline Gt2: An overview of an elephant in an enclosure. \newline Gt3: A small elephant stands alone in an enclosure. \newline Gt4: One elephant inside an enclosure at the zoo. \newline Gt5: An elephant in a fenced in enclosure packed with dirt.} \\

 \end{tabular}
  \label{tab:example_captions_1}
\end{table*}

\begin{table*}[htb]
 \centering
  \footnotesize
  \caption{Additional examples.}
  \begin{tabular}{ p{0.3cm} @{}c@{} | >{\RaggedRight\arraybackslash}p{4.8cm} | >{\RaggedRight\arraybackslash}p{4.8cm} } 
 {} & Image & Networks prediction & Ground Truth \\
 \hline
 5 &
 \raisebox{-0.9\totalheight}{
 \includegraphics[height=0.3\textwidth,width=0.3\textwidth]{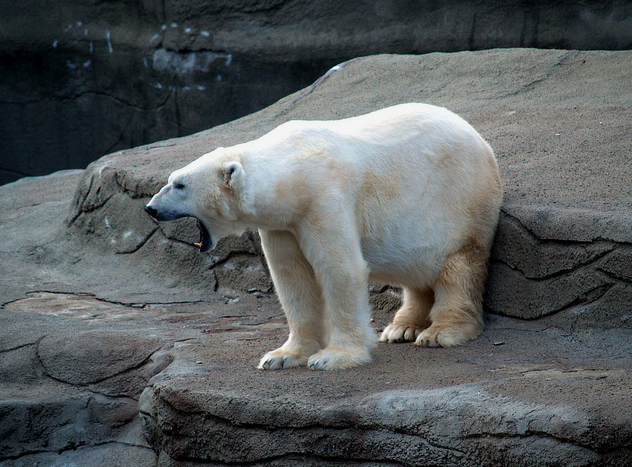}}
& {Base: A polar bear with a toy on top of it. \leavevmode\newline \leavevmode\newline
Base w/ AoA: A polar bear holding a man in its mouth. \leavevmode\newline \leavevmode\newline
ExpansionNet\textsubscript{\raisebox{-1pt}{\rlap{64,16}}}\textsuperscript{\raisebox{1pt}{Ablation}}: A polar bear walking on top of a rock. \leavevmode\newline \leavevmode\newline
ExpansionNet\textsubscript{64,16}: A polar bear is standing on top of a rock.
} & {Gt1: A close up of a polar bear on a rock formation. \newline Gt2: A polar bear yawning while standing on a rock. \newline Gt3: A polar bear opening its mouth while standing on a rock. \newline Gt4: A polar bear standing on a large rock with its mouth wide open. \newline Gt5: A large polar bear stands on rock with an open mouth.} \\

 \hline
 6 & 
 \raisebox{-0.9\totalheight}{
 \includegraphics[height=0.3\textwidth,width=0.3\textwidth]{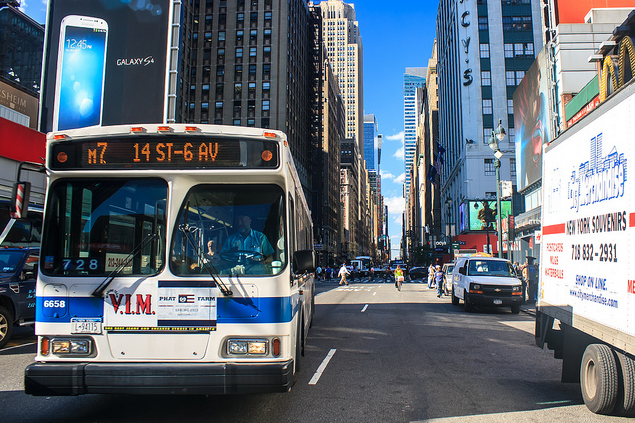}}
& {Base: A bus driving down a city street with a table. \leavevmode\newline \leavevmode\newline
Base w/ AoA: A bus dri\-ving down a city street with a man. \leavevmode\newline \leavevmode\newline
ExpansionNet\textsubscript{\raisebox{-1pt}{\rlap{64,16}}}\textsuperscript{\raisebox{1pt}{Ablation}}: A bus driving down a city street with cars.
\leavevmode\newline \leavevmode\newline
ExpansionNet\textsubscript{64,16}: A bus driving down a city street with buildings.
} & {Gt1: A bus and truck driving down a busy city street. \newline Gt2: A very busy city street with buses, vans, and semis. \newline Gt3: A white and blue bus on street with buildings in background. \newline Gt4: A city bus is going down a street. \newline Gt5: A transit bus riding down a busy city street.} \\

 \hline
 7 &
 \raisebox{-0.9\totalheight}{
 \includegraphics[height=0.3\textwidth,width=0.3\textwidth]{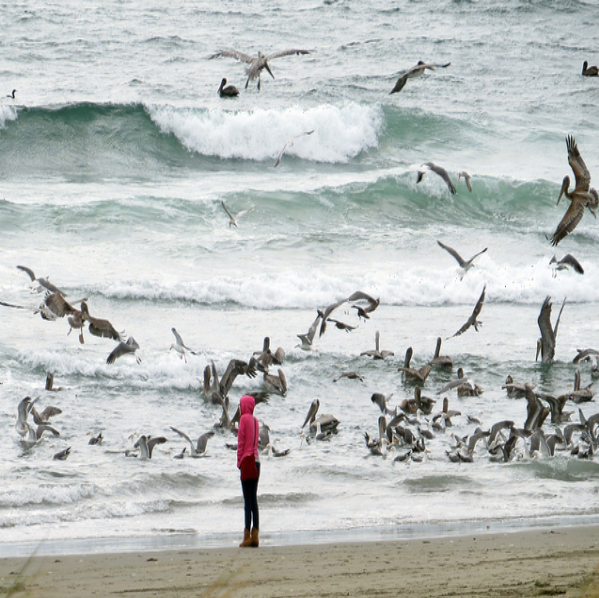}}
& {
Base: A woman standing on a beach near the water. 
\leavevmode\newline \leavevmode\newline Base w/ AoA: A woman standing in the water with a flock of birds. 
\leavevmode\newline \leavevmode\newline ExpansionNet\textsubscript{\raisebox{-1pt}{\rlap{64,16}}}\textsuperscript{\raisebox{1pt}{Ablation}}: 
A woman standing on a beach with a flock of birds.
\leavevmode\newline \leavevmode\newline
ExpansionNet\textsubscript{64,16}: A person standing on a beach with a flock of birds.
} & 
{Gt1: A woman standing on a beach is surrounded by birds. \newline
Gt2: Woman standing on a cold, pelican filled beach. 
\newline Gt3: A  woman standing on a sandy beach next to the ocean. \newline Gt4: A woman standing on a beach observing pelicans. \newline Gt5: A person is standing on a beach with a lot of birds.} \\

 \hline
 8 & 
 \raisebox{-0.9\totalheight}{
 \includegraphics[height=0.3\textwidth,width=0.3\textwidth]{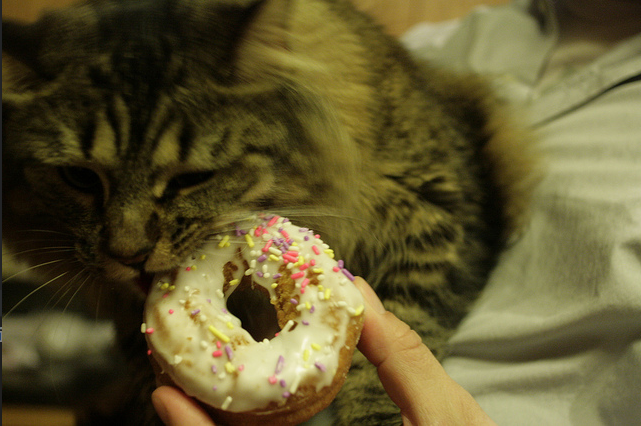}}
& {Base: A cat is eating a donut with a doughnut. \leavevmode\newline \leavevmode\newline
Base w/ AoA: A cat laying on a person holding a donut. \leavevmode\newline \leavevmode\newline
ExpansionNet\textsubscript{\raisebox{-1pt}{\rlap{64,16}}}\textsuperscript{\raisebox{1pt}{Ablation}}: A cat eating a donut with sprinkles.  \leavevmode\newline \leavevmode\newline
ExpansionNet\textsubscript{64,16}: A cat eating a donut with sprinkles.
} & {Gt1: A cat bites into a doughnut offered by a person's hand. \newline Gt2: A person is holding a doughnut up to a cat. \newline Gt3: A person feeding a donut with white frosting and sprinkles. \newline Gt4: A person is feeding a doughnut to a cat. \newline Gt5: A person holds a sprinkle covered doughnut to a cat's face.} \\

 \end{tabular}
  \label{tab:example_captions_2}
\end{table*} 

\section{Conclusions}

In this work we addressed the question of whether the input sequence length could pose a performance bottleneck in problems involving sequences. In order to do that, we presented a new method called expansion and introduced an architecture that exploits the advantages of working with different sequence lengths that change either statically or dynamically with respect to the input. Relying mostly on the expansion layers, with no attentive (beside the cross-connection) and recurrent parts, we conducted experiments on the MS-COCO 2014 captioning challenge and achieved a very competitive result, against the common belief that the latter approaches are required in order to perform well. In conclusion, we showed that networks are not necessarily bound to work with the input sequence length and by breaking such limitation we hope to lay the ground for a novel promising research direction with the goal of further improving currently existing models and new ones in the future.

\printbibliography[title=Bibliography]

\end{document}